%% file: conference_101719.tex
\documentclass[conference]{IEEEtran}
\IEEEoverridecommandlockouts
\usepackage{cite}
\usepackage{amsmath,amssymb,amsfonts}
\usepackage{algorithmic}
\usepackage{graphicx}
\usepackage{textcomp}
\usepackage{graphicx}
\usepackage{subfigure}
\usepackage{xcolor}
\def\BibTeX{{\rm B\kern-.05em{\sc i\kern-.025em b}\kern-.08em T\kern-.1667em\lower.7ex\hbox{E}\kern-.125emX}}

\begin{document}

\title{On the Applicability of Synthetic Data for Face Recognition \\
}
\author{Haoyu Zhang,  Marcel Grimmer, Raghavendra Ramachandra,  Kiran Raja,  Christoph Busch \\
Norwegian Biometrics Laboratory, Norwegian University of Science and Technology (NTNU), Norway\\  
	\{\tt\small haoyu.zhang; marceg; raghavendra.ramachandra;kiran.raja;christoph.busch\} @ntnu.no\\
	%	\{\tt\small l.j.spreeuwers;r.n.j.veldhuis@utwente.nl\}
}
% \and
% \and
% \and
% \and
% \and

\maketitle
\input{content/00-Abstract}

\input{content/01-Introduction}

\input{content/02-Methodology}

\input{content/04-Experiments}
\input{content/05-Conclusion}

\bibliographystyle{IEEEtran.bst}
\bibliography{references.bib}

\end{document}

%% file: content/00-Abstract.tex
\begin{abstract}

 Face verification has come into increasing focus in various applications including the European Entry/Exit System, which integrates face recognition mechanisms. At the same time, the rapid advancement of biometric authentication requires extensive performance tests in order to inhibit the discriminatory treatment of travellers due to their demographic background. However, the use of face images collected as part of border controls is restricted by the European General Data Protection Law to be processed for no other reason than its original purpose. Therefore, this paper investigates the suitability of synthetic face images generated with StyleGAN and StyleGAN2 to compensate for the urgent lack of publicly available large-scale test data. Specifically, two deep learning-based (SER-FIQ, FaceQnet v1) and one standard-based (ISO/IEC TR 29794-5) face image quality assessment algorithm is utilized to compare the applicability of synthetic face images compared to real face images extracted from the FRGC dataset. %In this context, the utility of a face image refers to the prediction of the biometric recognition accuracy. 
 Finally, based on the analysis of impostor score distributions and utility score distributions, our experiments reveal negligible differences between StyleGAN vs. StyleGAN2, and further also minor discrepancies compared to real face images.    

\end{abstract}


%% file: content/01-Introduction.tex
\section{Introduction}
\label{sec:introduction}

Biometric verification refers to the automated recognition of individuals based on their biological and behavioural characteristics \cite{ISO_2382_37}. Among multiple biometric characteristics, face has gained popularity in various application scenarios, such as border control systems like the European Entry-Exit System (EES) \cite{EU-Regulation-EES-InternalDocument-2017}, passport issuance and civilian ID management. Human face as a biometric modality has proven to be sufficiently unique to allow individual recognition with reasonable inter-class distance. Driven by the factors such as user convenience and good biometric performance (i.e., both identification and verification), the Smart Borders program under EES initiative has identified face as a mandatory biometric data. The EES system will be deployed as a central system for collecting and querying traveller data including face data to the Schengen area at all border crossing points\cite{EU-Regulation-EES-InternalDocument-2017}. %The facial images are captured for authentication due to the high user  convenience and efficiency associated with the data acquisition process.

The deployment of biometric recognition at the European borders requires the biometric performance to comply with high standards defined in the best practices for automated border control of the European Border and Coast Guard Agency (Frontex) \cite{FRONTEX-BorderControl-BestPractices-InternalDocument-2015}. While the value of real and sufficiently representative biometric data for biometric performance tests is obvious, it is becoming more and more important that the business can reasonably consider also alternative options such as the
generation and use of synthetic biometric data samples, ensuring comparable characteristics and representativeness to real data. However, generating synthetic face images with similar properties to real face images captured
at border control scenarios (e.g. frontal head poses without
face occlusions) continues to be a technical challenge.

\begin{figure}[t]
    \centering
    \includegraphics[width=0.75\linewidth]{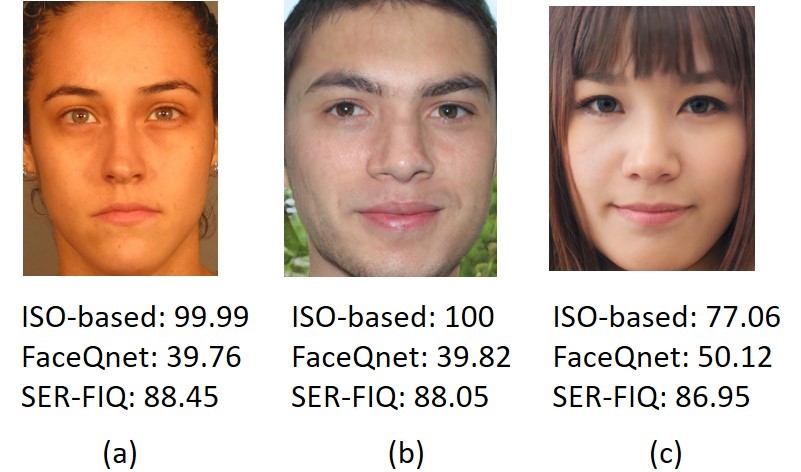}
    \caption{Example face images from different datasets annotated with three different FQAAs. (a) FRGC (b) StyleGAN (c) StyleGAN2}
    \label{fig:qs-examples}
\end{figure}

In an attempt to answer this question, this work analyses the differences between synthetically generated face images and real face images from the biometric perspective. Although biometric performance reporting can be a way to assess the applicability of the synthetically generated data, the random generation procedure often produces limited true-mated pairs. Reporting the verification performance on synthetically generated dataset is therefore not realistic.  Among other approaches for conducting such an analysis, we choose to assess the quality of the face images reflecting regular biometric systems where quality is first determined before using it for any biometric purpose.  Face Quality Assessment Algorithms (FQAA) quantify the biometric quality of a sample (for instance, face image) by translating it to a quality score between [0, 100] \cite{ISO_29794_1:2016}. A high quality score indicates that the corresponding biometric sample is well suited for biometric recognition and a low quality score leads to inaccurate results due to the low quality of the input image. This understanding of biometric quality complies with the terminology of ISO/IEC 29794-1 \cite{ISO_29794_1:2016}, which defines the utility of a biometric sample as the prediction of the biometric recognition accuracy. In general, FQAAs are either aiming to predict the utility of face images for a specific face recognition model or alternatively try to establish a general utility predictor that can be used for an arbitrary face recognition system. In this context, Terhörst et al.~\cite{terhorst2020ser} introduced a model-specific FQAA, based on the stochastic embedding robustness ("SER-FIQ") of quality features extracted with ArcFace \cite{deng2019arcface}. The comparison of the authors against previous state-of-the-art FQAA approaches showed that SER-FIQ significantly outperformed alternative methods. Recently, Hernandez-Ortega \cite{Hernandezortega-FQA-FaceQnetV1-2020} presented FaceQnet v1, a general FQAA, the performance of which was benchmarked in the ongoing quality assessment evaluation of the National Institute of Standards and Quality (NIST) \cite{Grother-FRVT-Identification-NISTIR-2020}. However, despite state-of-the-art performances of SER-FIQ and FaceQnet v1, the utility predictions of both algorithms are not explainable. For this reason, face image quality features based on a technical report of international standardisation bodies, such as ISO/IEC TR 29794-5 \cite{ISO_29794_5}, can be utilized to manually construct a more explainable FQAA, as proposed by Wasnik et al.~\cite{wasnik2017assessing} (see FFigure \ref{fig:qs_iso_realvssyn}).

By measuring the face quality metrics of synthetically generated face samples, we can establish if they can be used for biometric algorithm training and testing purposes. In order to assert and evaluate our hypothesis, we conduct an extensive analysis of three different quality metrics such as ISO/IEC TR 29794-5 \cite{ISO_29794_5} and two deep learning based quality metrics \cite{Hernandezortega-FQA-FaceQnetV1-2020, terhorst2020ser} on face images generated using StyleGAN \cite{karras2019style} and StyleGAN2 \cite{karras2020analyzing} respectively. StyleGAN2 emphasizes the capability of the new architecture to generate higher quality face images by removing water droplet-like artefacts caused by StyleGAN. However, the question of whether StyleGAN2 images are more suitable for face recognition compared to StyleGAN images has not yet been addressed. Therefore, this paper also includes a comparison between StyleGAN and StyleGAN2 based on recent face image quality assessment algorithms (FQAAs). Figure \ref{fig:qs-examples} illustrates the quality scores obtained on face images from different datasets. As noted from Figure \ref{fig:qs-examples}, various FQAAs indicate different face image quality score. Noting that they are inconsistent in providing similar scores, we evaluate the synthetically generated face images using all three FQAAs to assess the usability of such images in biometrics from different complementary perspectives.

The main contribution of this paper is in investigating and reporting to which extent the synthetic samples could replace the need for real images for conducting biometric performance tests of a pre-trained face recognition system. More specifically, the goal is to obtain synthetic face images that are similar to real face images not only from the human perception point, but also in the context of face recognition. In this paper, the differences between synthetic and real face images in terms of their utility are evaluated by measuring the quality with two deep learning based FQAAs: FaceQnet v1 \cite{Hernandezortega-FQA-FaceQnetV1-2020} and SER-FIQ \cite{terhorst2020ser}. Additionally, based on the work of Wasnik et al. \cite{wasnik2017assessing}, a random forest regressor is trained on manually extracted quality features defined by ISO/IEC TR 29794-5:2010 \cite{ISO_29794_5}.

%% file: content/02-Methodology.tex
\section{Synthetic Face Image Generation and Dataset}
\label{sec:methods}

%\subsection{StyleGAN and StyleGAN2}
 Karras et al. \cite{karras2019style} presented a style-based generator architecture for generative adversarial networks (StyleGAN), capable of generating synthetic images with high-resolution (1024x1024) and realistic appearances. StyleGAN applies latent space mapping and adaptive instance normalization (AdaIN) to control the image synthesis process by styles, which finally leads to the unsupervised separation of high-level attributes. In addition to their proposed GAN architecture, the authors webcrawled high-quality human face images from a social media platform (Flickr) in order to create a new dataset (FFHQ), which covers a wide variation of soft biometrics (e.g. ethnicity, gender). Extending the work of StyleGAN, Karras et al \cite{karras2020analyzing} further published the state-of-the-art model StyleGAN2, where they improved the architectural design and fixed the characteristic artefacts occurring in the synthetic images generated by StyleGAN. 
 
\input{content/03-Dataset}

\section{Face Quality Assessment Algorithms}
\label{sec:fqaa}
For the convenience of the reader, we provide a brief background of various FQAA. Our rationale for choosing them for the evaluations is to cover a standardized algorithm together with deep learning algorithms in the emerging direction. The range of the output scores in these algorithms is originally [0,1] but has been scaled to [0,100] following \cite{ISO_29794_1:2016}.
\subsection{ISO/IEC TR 29794-5:2010 Implementation}
ISO/IEC 29794-5:2010 \cite{ISO_29794_5} Implementation uses the hand-crafted quality metrics and is based on the work of Wasnik \cite{wasnik2017assessing}. For each image under assessment, various hand-crafted quality metrics for facial images following the technical report ISO/IEC TR 29794-5 are extracted as a feature vector. Given the feature vector as an input, then a pre-trained Random Forest Regressor is applied to predict a quality score. As against other deep learning models, this approach ensures a better explainability basing the results on various hand-crafted features.

\subsection{FaceQnet v1}

FaceQnet v1 is a deep learning based FQAA proposed by Hernandez-Ortega et al. \cite{hernandez2020biometric} and aims to predict the general utility of a face image, independent from a specific face recognition system. For the quality score prediction, a pre-trained network of ResNet-50 is fine-tuned on a small subset of VGGFace2 dataset \cite{cao2018vggface2} including 300 data subjects. All images contained in VGGFace2 represent webcrawled celebrities with a large variation in pose, age, illumination, etc. Since ResNet-50 is already trained for face recognition, the authors assume that the same network weights can be exploited for assessing the quality of a face image.

FaceQnet v1 follows a supervised learning approach, which means that the ground truth quality scores are required for fine-tuning the model. Finding representative quality scores that accurately reflect general utility criteria is a challenging task. Therefore, the authors assume if an ICAO 9303 \cite{ICAO9303} compliant image A represents perfect image quality, the mated comparison score between image A (ICAO compliant) and image B (unknown quality) measures the utility of image B. If the mated comparison score is low, this must be due to image B, since image A fulfills the ICAO quality criteria. On the other hand, if the comparison score is high, it can be assumed that also image B is of good quality.

The authors employed the BioLab-ICAO framework \cite{ferrara2012face} to select high-quality images per subject, which are used as a reference for computing the ground truth quality score for the remaining training images. The ground truth quality scores are obtained by calculating and fusing the comparison scores of mated samples, which represents the utility of the non-ICAO compliant sample.

\subsection{SER-FIQ}

Recently, Terhörst et al \cite{terhorst2020ser} introduced an estimation method for predicting the face image quality based on stochastic embedding robustness (SER-FIQ). SER-FIQ is an unsupervised technique that is not dependent on previously extracted ground truths in order to train the prediction model. Compared to FaceQnet v1, which outputs the general utility of a face image, SER-FIQ focuses on predicting the utility of a specific face recognition system. More precisely, the quality scores are based on the variations of face embeddings stemming from random subnetworks of a face recognition model. The authors argue that a high variation between the embeddings of the same sample functions as a robustness indication, which is assumed to be synonymous with image quality.

Since the face embeddings extracted with face recognition systems are deterministic, the main idea of SER-FIQ is to add dropout layers as additional components to create random subnetworks for each prediction of the same sample. Once a fixed number of stochastic embeddings are extracted, the sigmoid of the negative mean euclidean distances between all embedding pairs is computed and outputs a quality score. However, the computational complexity of SER-FIQ increases quadratically with the number of random subnetworks, which leads to a trade-off between the efficiency of the algorithm and the expected accuracy of the quality predictions. Following the recommendation of the authors, all experiments in this report are conducted with a number of N=100 subnetworks.

% Empirically, we noticed that the original implementation is not sufficiently sensitive to the variation of illumination condition. Hence, we additionally introduced two metrics:

% Another metric of light symmetry measures the histogram of the left half and the right half by Chi-Square (instead of Gabor features and L1 norm in the existing implementation).
% Illumination uniformity metric measures the difference between signal-to-noise ratio of the left half and the right half of the image.
% Compared with previous algorithms,  this implementation uses hand-crafted features, which ensures a better explainability and less training time but also may lead to limited generalizability than models based on deep learning. As for efficiency,  processing the images by hand-crafted metrics is slower than end-to-end inference in FaceQnet but faster than SER-FIQ.

%% file: content/03-Dataset.tex
\subsection{Dataset}
\label{sec:dataset}
%In this section, we describe the constitution of our databases for face image quality assessment. 
In order to assert our hypothesis, we employ a synthetically generated face image dataset.  Our synthetic data is generated using StyleGAN \cite{karras2019style} and StyleGAN2 \cite{karras2020analyzing} models which are pre-trained on the FFHQ \cite{karras2019style} dataset. These generated images are of resolution $1024 \times 1024$ pixels with high visual quality.
The synthetic image generation by sampling latent codes may result in images choosing extreme regions of the latent space causing unwanted variation. To avoid such an adverse effect and mitigate images of unwanted visual appearance, we apply a truncation factor on the latent space of both StyleGAN and StyleGAN2. The truncated latent code $w'$ can be represented as $w' = \Bar{w} + \psi (w-\Bar{w})$
 and is generated by scaling the deviation between the original latent code $w$ and $\Bar{w}$, the center of mass of the latent space, with a truncation factor $\psi \leq 1$. The truncation factor therefore stabilizes the image generation by avoiding sampling latent codes in extreme regions of the latent space but will also cause some loss of variation as a trade-off. To provide an insight on the impact of the truncation factor on generated biometric quality, we also include sub-datasets generated with truncation factors varying from 0.25, 0.5 and 0.75 where each of the configurations has 50,000 random samples.   

\begin{figure}
    \centering
    \includegraphics[width=1\linewidth]{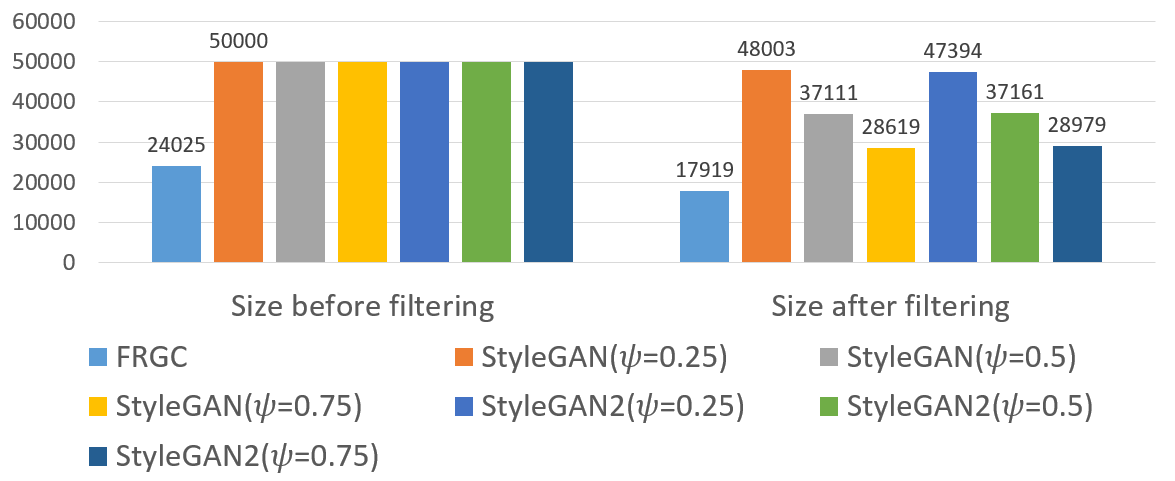}
    \caption{Constitution and origin of our datasets}
    \label{fig:dataset}
\end{figure}

To compare the synthetic data with real face data, we chose a subset containing 24,025 images from the FRGC-V2 face database \cite{FRGC_DB} as our representative dataset due to relatively high-quality images and constrained conditions that resemble the image quality in EES cases.
To ensure that our synthetic datasets have comparable conditions, we also discard some unsatisfying images by having pre-selection criteria such as minimum inter-eye distance (IED), illumination metrics and predicted head poses to further improve the consistency between our assessment and the facial quality requirements proposed by ISO/IEC TR 29794-5:2010 \cite{ISO_29794_5}, ICAO 9303 \cite{ICAO9303}. The key motivation is also to mimic automatic border control, one of the most common and important real-life scenarios of FRS. The statistical distribution information of our datasets is presented in Figure~\ref{fig:dataset}.
% More specifically, we use a pre-trained landmark detection model to predict the center of eyes and filter out facial images with IED $\leq$ 90 pixels. The detection process also filters out images with covered eyes. For illumination, we prepared a dataset of positive samples with good face quality and negative samples captured in the poor illumination condition. Hand-crafted features are extracted and then a Support Vector Machine is trained for classification. To filter out images with rotated head poses, we use img2pose model [17] to predict and filter out images with yaw angles beyond [-15°, 15°] , pitch angles beyond  [ -25°, 20°] and roll angles beyond  [-20°, 20°].

%% file: content/04-Experiments.tex
\section{Experiments and Results}
\label{sec:experiments}

In this section, the experimental details for assessing the quality of synthetic data are introduced and the corresponding quantitative and qualitative results are discussed. We first provide the differences between synthetic datasets generated with StyleGAN and StyleGAN2, taking into account various truncation factors. Based on these results, we also present the analysis on the distinction between synthetic and real face images.

In order to compare the utility of face images from different datasets, impostor distributions of comparison scores are created to evaluate differences in the similarity among non-mated face images. The impostor distribution can show the diversity of identity in each dataset and also can evaluate its verification performance on FRS, in the circumstance that our synthetic data is randomly generated without any mated samples. In the impostor distribution, 5,000 random pairs are randomly sampled from the dataset and the cosine similarities between their embeddings extracted with Arcface \cite{deng2019arcface} are calculated. Then, comparison scores of each dataset are converted to histograms and fitted to a Gaussian distribution using kernel density estimation. As a reference, the threshold=0.25 of the applied Arcface model is set at False Match Rate (FMR) = 0.1\% on LFW dataset \cite{huang2008labeled} following the guidelines of Frontex \cite{Frontex}. In addition to the impostor scores, we apply two deep learning-based and one standard FQAA to predict the utility of the face images. Once the quality scores are extracted, the differences between histograms of different datasets are evaluated using Kullback-Leibler divergence which measures how one distribution is different from another.

\subsection{Comparison between StyleGAN and StyleGAN2}
As introduced in Section \ref{sec:introduction}, Karras et al \cite{karras2020analyzing} have stated that StyleGAN2 has achieved an improvement in image quality. However, it is still necessary to further evaluate the difference of generated data between StyleGAN and StyleGAN2 from the perspective of biometric data instead of general images.

\begin{figure*}[htbp]
\centering
\subfigure[KL-D: 0.702]{
\begin{minipage}[t]{0.31\linewidth}
\centering
\includegraphics[width=2.1in]{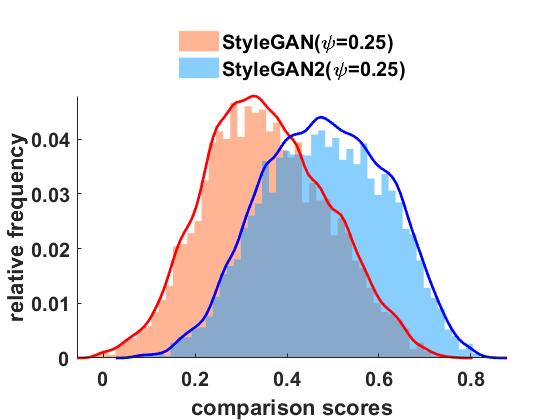}
%\caption{fig1}
\end{minipage}%
}%
\subfigure[KL-D: 0.107]{
\begin{minipage}[t]{0.31\linewidth}
\centering
\includegraphics[width=2.1in]{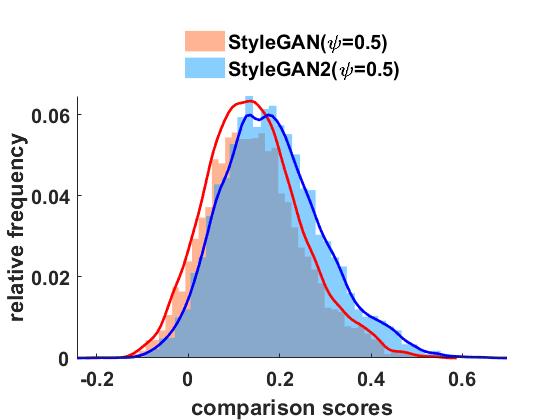}
%\caption{fig2}
\end{minipage}%
}%
\subfigure[KL-D: 0.007]{
\begin{minipage}[t]{0.31\linewidth}
\centering
\includegraphics[width=2.1in]{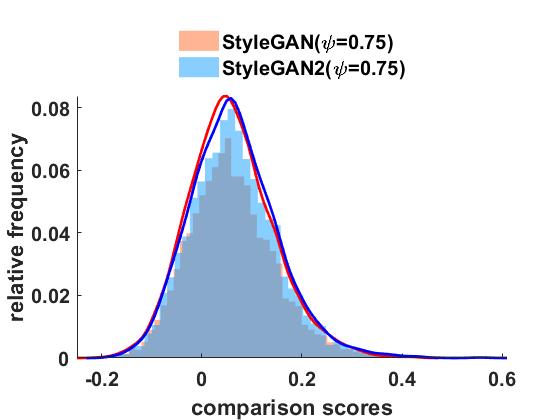}
%\caption{fig2}
\end{minipage}
}%
\centering
\caption{Comparison of impostor distributions between StyleGAN and StyleGAN2 using Arcface \cite{deng2019arcface} (threshold=0.25 @ FMR=0.1\% on LFW \cite{huang2008labeled} dataset). (a) FaceQnet v1 (b) Random Forest Regressor (ISO/IEC TR 29794-5) (c) SER-FIQ }
\label{fig:imposter_ivs2}
\end{figure*}

Figure \ref{fig:imposter_ivs2} presents impostor distributions of StyleGAN and StyleGAN2. It is shown that the dataset generated with StyleGAN2 using a truncation factor of $\psi=0.25$ has a higher mean value of comparison score than StyleGAN. However, this difference is reducing with the increase of $\psi$ and vanishes as the truncation factor increases to 0.75. This also fits our original understanding of the truncation that it shrinks the sampling region of the latent space so the generated facial images will have less variation of conditions. However, this may also raise the concern that the generated samples will lack a diversity of identity information. As shown in Figure \ref{fig:imposter_ivs2}, it is obvious that reducing the truncation factor will lead to an increase in the mean value of comparison scores, which means that the non-mated samples tend to have more similar identity information.  

\begin{figure*}[htbp]
\centering
\subfigure[KL-D: 0.008]{
\begin{minipage}[t]{0.31\linewidth}
\centering
\includegraphics[width=2.2in]{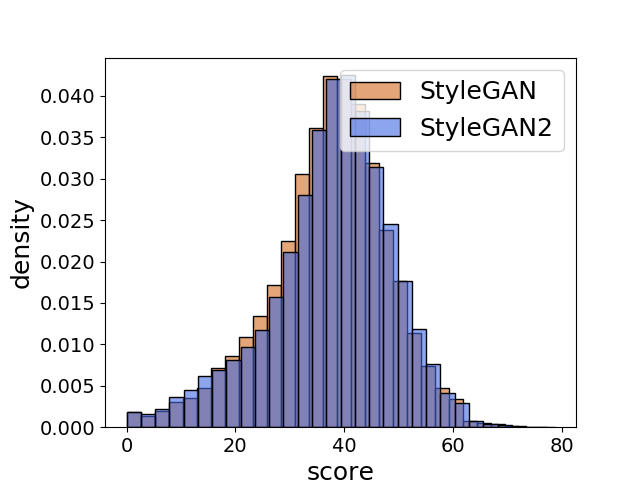}
%\caption{fig1}
\end{minipage}%
}%
\subfigure[KL-D: 0.003]{
\begin{minipage}[t]{0.31\linewidth}
\centering
\includegraphics[width=2.2in]{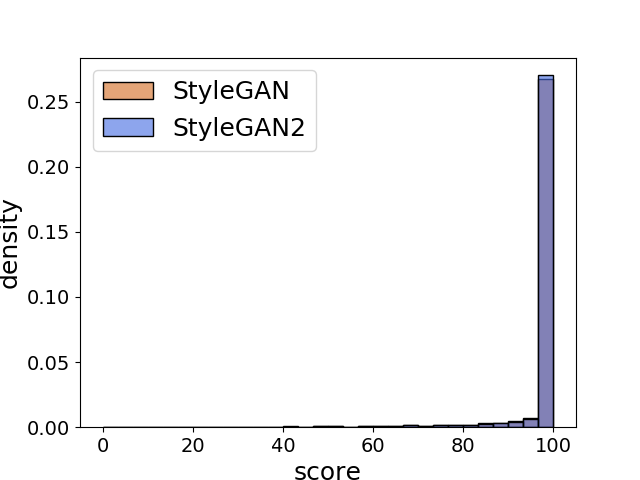}
%\caption{fig2}
\end{minipage}%
}%
\subfigure[KL-D: 0.116]{
\begin{minipage}[t]{0.31\linewidth}
\centering
\includegraphics[width=2.2in]{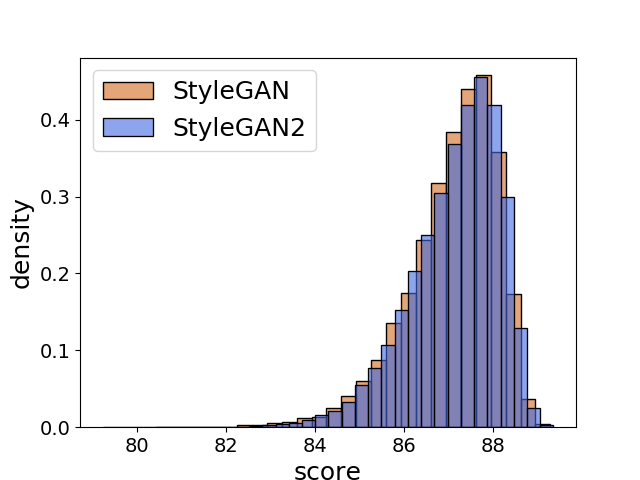}
%\caption{fig2}
\end{minipage}
}%
\centering
\caption{Comparing the quality score distributions from various face quality algorithms between StyleGAN and StyleGAN2. (a) FaceQnet v1 (b) Random Forest Regressor (ISO/IEC TR 29794-5) (c) SER-FIQ }
\label{fig:qs_ivs2}
\end{figure*}

To further validate the findings based on the impostor scores, the distributions of the face quality scores from different FQAAs are shown in Figure \ref{fig:qs_ivs2}, limited to a single truncation factor of $\psi=0.75$. Similar to the impostor scores, the distributions between StyleGAN and StyleGAN2 are close to identical across all FQAAs, thus reinforcing the conclusion that StyleGAN and StyleGAN2 face images are equally suitable for biometric recognition. 

\subsection{Comparison between Synthetic and Real Data}

\begin{figure}
    \centering
    \includegraphics[width=0.58\linewidth]{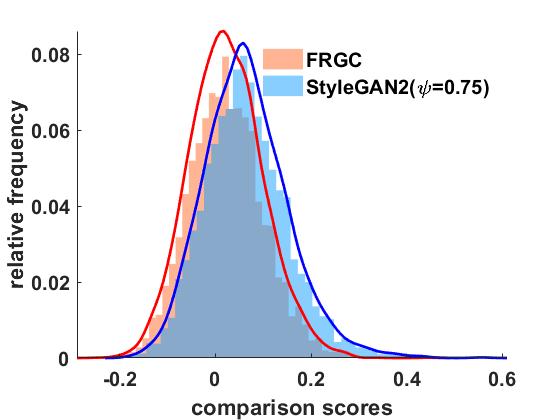}
    \caption{Impostor comparison score distributions of randomly selected StyleGAN and FRGC image pairs using Arcface \cite{deng2019arcface} (threshold=0.25 @ FMR=0.1\% on LFW \cite{huang2008labeled} dataset). KL-D: 0.184.}
    \label{fig:imposter_realvssyn}
\end{figure}

\begin{figure*}[htbp]
%\centering
%\subfigure[]{
%\begin{minipage}[t]{0.31\linewidth}
%\centering
%\includegraphics[width=2.45in]{Figures/qs_realvssyn/faceqnet_fvs1.png}
%\caption{fig1}
%\end{minipage}%
%}%
%\subfigure[]{
%\begin{minipage}[t]{0.31\linewidth}
%\centering
%\includegraphics[width=2.45in]{Figures/qs_realvssyn/ISO_fvs1.png}
%\caption{fig2}
%\end{minipage}%
%}%
%\subfigure[]{
%\begin{minipage}[t]{0.31\linewidth}
%\centering
%\includegraphics[width=2.45in]{Figures/qs_realvssyn/serfiq_fvs1.png}
%\caption{fig2}
%\end{minipage}
%}%

\centering
\subfigure[KL-D: 0.111]{
\begin{minipage}[t]{0.28\linewidth}
\centering
\includegraphics[width=2.1in]{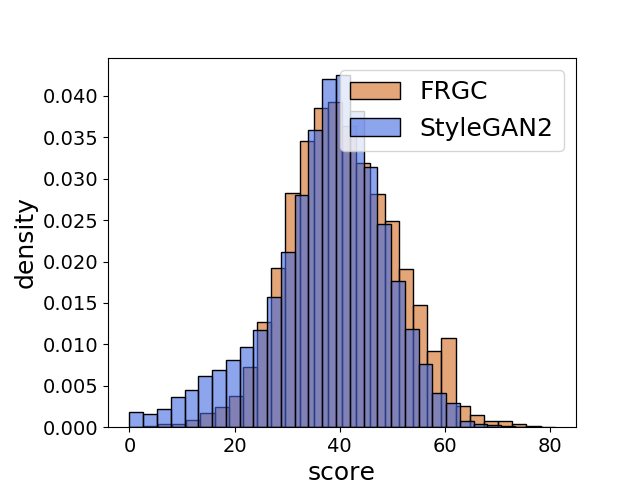}
%\caption{fig1}
\end{minipage}%
}%
\subfigure[KL-D: 0.444]{
\begin{minipage}[t]{0.28\linewidth}
\centering
\includegraphics[width=2.1in]{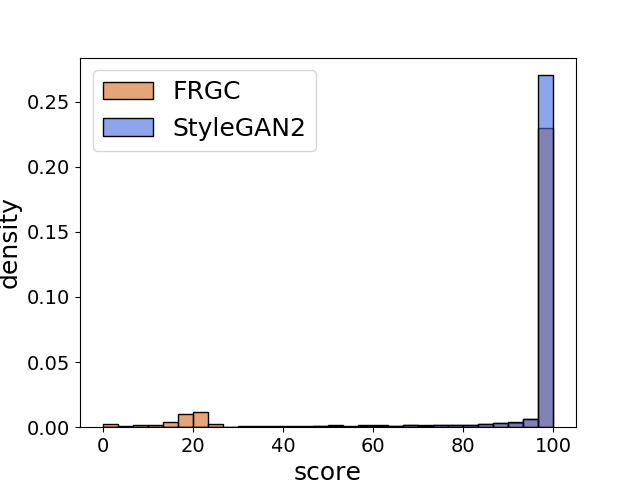}
%\caption{fig2}
\end{minipage}%
}%
\subfigure[KL-D: 1.172]{
\begin{minipage}[t]{0.28\linewidth}
\centering
\includegraphics[width=2.1in]{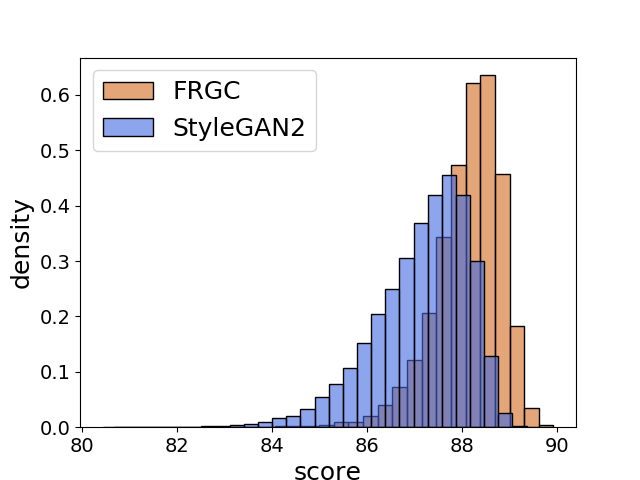}
%\caption{fig2}
\end{minipage}
}%
\centering
\caption{Comparing the quality score distributions from various face quality algorithms between FRGC and StyleGAN2. (a) FaceQnet v1 (b) Random Forest Regressor (ISO/IEC TR 29794-5) (c) SER-FIQ }
\label{fig:qs_realvssyn}
\end{figure*}

% \begin{figure*}
%     \centering
%     \includegraphics[width=0.85\linewidth]{Figures/qs_realvssyn.jpg}
%     \caption{Comparison of quality score distributions between FRGC and StyleGAN ($\psi= 0.7$). Left:  FaceQnet v1 (KL-D: 0.17), Right: SER-FIQ (KL-D: 1.05).}
%     \label{fig:qs_realvssyn}
% \end{figure*}

\begin{figure*}[htbp]
%\centering
%\subfigure[]{
%\begin{minipage}[t]{0.55\linewidth}
%\centering
%\includegraphics[width=4in]{Figures/qs_realvssyn/iso_features_realvs1.png}
%\end{minipage}%
%}%
%\subfigure[]{
%\begin{minipage}[t]{0.45\linewidth}
%\centering
%\includegraphics[width=3in]{Figures/qs_realvssyn/bluriness_realvs1.png}
%\caption{fig2}
%\end{minipage}%
%}%
\centering
\subfigure[]{
\begin{minipage}[t]{0.41\linewidth}
\centering
\includegraphics[width=2.6in]{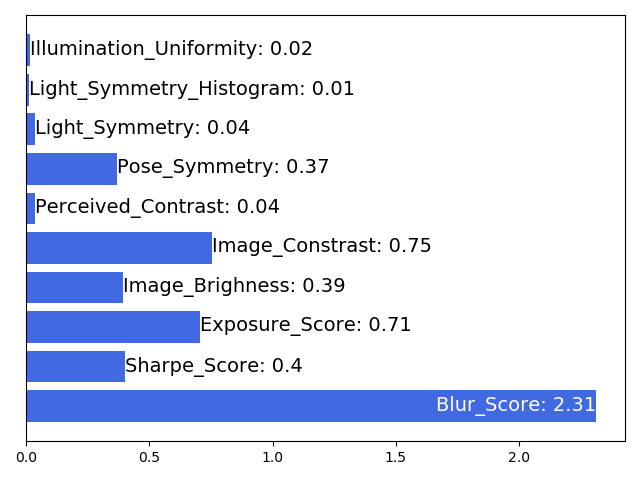}
\end{minipage}%
}%
\subfigure[]{
\begin{minipage}[t]{0.41\linewidth}
\centering
\includegraphics[width=2.8in]{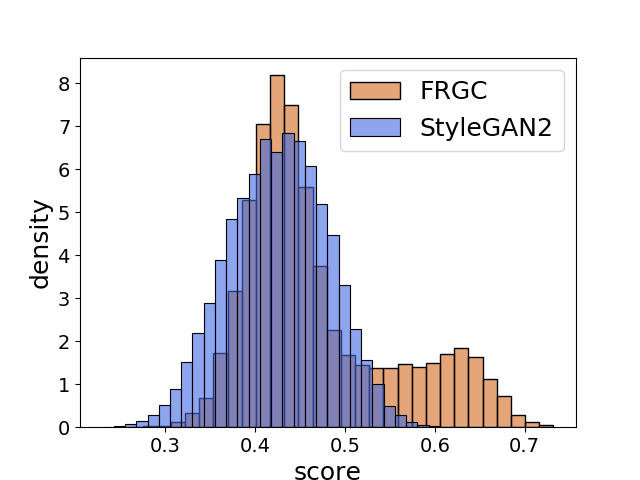}
%\caption{fig2}
\end{minipage}%
}%
\centering
\caption{Comparing the quality features from ISO/IEC TR 29794-5:2010 implementation between FRGC and StyleGAN2 ($\psi= 0.75$). (a) Kullback-Leibler Divergences between each quality features (b) Comparison of bluriness score distributions}
\label{fig:qs_iso_realvssyn}
\end{figure*}

In the previous section, the comparison between StyleGAN and StyleGAN2 has revealed that as the truncation factor increases, the differences between the generated images vanish. Meanwhile, it has been shown that the variation of identity information of generated images is limited when the truncation factor is small. Based on this observation, this section focuses on the comparison of the utility between synthetic and real face images. More precisely, the synthetic images are generated with StyleGAN2 ($\psi=0.75$) and compared to real face images from the publicly available FRGC dataset (see section \ref{sec:dataset}).

Figure \ref{fig:dataset} shows the proportion of filtered out images: While only 58\% of the StyleGAN2 images passed the filtering pipeline, more FRGC images could be retained with a rate of 75\%. The huge difference can be explained by the way the data samples were acquired: StyleGAN2 is trained on the FFHQ dataset, which has been webcrawled from a social media platform (Flickr). On the other hand, the images contained in FRGC were physically captured to use them for face recognition testing. Therefore, the variability of the StyleGAN2 images is much higher compared to those included in FRGC. In the same context, the number of filtered out images due to extreme pose rotations turns out to be much lower for FRGC in contrast to StyleGAN2 images. 

Analogous to the last subsection, Figure \ref{fig:imposter_realvssyn} compares the impostor score distributions between StyleGAN2 and FRGC non-mated pairs. Since we are investigating whether real face images can be replaced by synthetic ones, the degree of overlapping areas is of particular interest. In this context, the inspection of Figure \ref{fig:imposter_realvssyn} shows that both distributions have a similar gaussian-curved shape with huge overlapping areas. However, it is also visible that the right tail of the StyleGAN2 distribution is heavier compared to the FRGC dataset. In other words, the similarity scores of non-mated comparisons are slightly higher compared to those of FRGC, which potentially causes higher False-Match-Rates. Nevertheless, the minor differences might also be due to random effects rooted in the data acquisition process.

To further evaluate the discrepancies between StyleGAN2 and FRGC, Figure \ref{fig:qs_realvssyn} depicts the quality score distributions for different FQAAs. Looking at the FaceQnet v1 distributions, both areas are nearly identical with a very low Kullback Leibler Divergence of 0.111. However, steering the focus on the SER-FIQ distributions, a clear shift in the peaks is notable, which reveals that the estimated utility of the synthetic images is lower compared to FRGC images. This behavior corresponds to our expectations as the variety of pose rotations is much higher within the StyleGAN2 images and the authors highlight the sensitivity of their approach to this kind of deviation.

Finally, Figure \ref{fig:qs_iso_realvssyn} shows the Kullback-Leibler divergences between distributions of the individual ISO/IEC TR 29794-5:2010 quality features. At first glance, the blurriness score stands out, which indicates a significant difference between the real and synthetic datasets. A deeper look at the two distributions in Figure \ref{fig:qs_iso_realvssyn}(b) reveals that several FRGC images are annotated with high blurriness scores, which manifests itself in a bimodal distribution. The reason for this observation is due to the capturing process, where the images of multiple subjects were captured in motion.

%% file: content/05-Conclusion.tex
\section{Conclusion}
\label{sec:conclusion}
To investigate the suitability of synthetically generated face images for biometric recognition, we evaluated the face quality of synthetic face images generated by StyleGAN and StyleGAN2 and also compared to real face images from the FRGC dataset which represents the quality of images in border control. The comparison of the utility between different datasets is based on histograms of non-mated comparison scores with Arcface \cite{deng2019arcface} and histograms of quality scores with 3 different representative face quality algorithms, including the state-of-the-art SER-FIQ \cite{terhorst2020ser} algorithm. 

The first part of our experiments focuses on the comparison between different synthetic face datasets. The evaluation result of comparing synthetic face data generated by StyleGAN and StyleGAN2 shows that their utility for biometric recognition is similar. In this context, both the impostor comparison scores and the quality scores are consistent in their results and indicate that as the truncation factor increases, the differences vanish. By comparing various truncation factors, it is also demonstrated that while the image quality increases with smaller truncation values, the variety of the generated face images decreases at the same time. In other words, a low truncation factor leads to face images with similar identities and is therefore not suitable for biometric performance in border control scenarios, such as the EES.

The second part of the experiments is focusing on the comparison between the synthetic and real face images. The impostor distribution of synthetic data is slightly offset to the right, which means that the similarity between non-mated samples is slightly higher compared to real face images. The analysis of the FaceQnet v1 quality scores has not revealed any major differences between images stemming from StyleGAN2 or FRGC. However, the comparison of the SER-FIQ quality scores revealed a minor drop in quality for synthetic face images, which can be explained by the wider range of pose rotations. Further, the assessment of the ISO/IEC TR 29794-5:2010 features has shown a high blurriness score for FRGC images, caused by capturing some face images in motion. Finally, we conclude that StyleGAN2 and FRGC images have shown the minor differences in face quality, which means the evaluated synthetic data can achieve a similar quality as biometric samples in EES cases and allows us to exploit both domains for realistic biometric performance tests.